# Inference in Hybrid Bayesian Networks Using Mixtures of Gaussians

**Prakash P. Shenoy**
University of Kansas School of Business
1300 Sunnyside Ave, Summerfield Hall
Lawrence, KS 66045-7585 USA
pshenoy@ku.edu

## Abstract

The main goal of this paper is to describe a method for exact inference in general hybrid Bayesian networks (BNs) (with a mixture of discrete and continuous chance variables). Our method consists of approximating general hybrid Bayesian networks by a mixture of Gaussians (MoG) BNs. There exists a fast algorithm by Lauritzen-Jensen (LJ) for making exact inferences in MoG Bayesian networks, and there exists a commercial implementation of this algorithm. However, this algorithm can only be used for MoG BNs. Some limitations of such networks are as follows. All continuous chance variables must have conditional linear Gaussian distributions, and discrete chance nodes cannot have continuous parents. The methods described in this paper will enable us to use the LJ algorithm for a bigger class of hybrid Bayesian networks. This includes networks with continuous chance nodes with non-Gaussian distributions, networks with no restrictions on the topology of discrete and continuous variables, networks with conditionally deterministic variables that are a nonlinear function of their continuous parents, and networks with continuous chance variables whose variances are functions of their parents.

## 1 INTRODUCTION

This paper is about making exact inferences in hybrid Bayesian networks, i.e., Bayesian networks with a mix of discrete, continuous, and conditionally deterministic variables. One of the earliest algorithms for this class of problems was proposed by Lauritzen [1992], and later revised by Lauritzen and Jensen [2001]. We refer to this algorithm as the *LJ algorithm*. The LJ algorithm applies only to Bayesian networks where the continuous variables have the so-called conditional linear Gaussian distributions, and to networks such that discrete variables do not have continuous parents. In this case, for every instantiation of the discrete variables, the joint distribution of the continuous variables is multivariate Gaussian. Thus, we will refer to such Bayesian networks as mixtures of Gaussians (MoG) Bayesian networks. The LJ algorithm is implemented in Hugin, a commercially available software.

One approach to exact inference in hybrid Bayesian networks is suggested by Murphy [1999] and Lerner *et al.* [2001]. This approach is only applicable to so-called "augmented CLG networks," which are hybrid Bayesian networks with conditional linear Gaussian distributions for continuous variables, and which allow discrete variables with continuous parents. The main idea of this approach is to approximate the product of a Gaussian and a logistic distribution (for discrete variables with continuous parents) by a variational approximation [Murphy 1999] or by mixture of Gaussians using numerical integration [Lerner *et al.* 2001]. This idea is then embedded within the general framework of the LJ algorithm.

Another approach to exact inference in hybrid Bayesian networks is based on the idea by Moral *et al.* [2001] of using mixtures of truncated exponentials to approximate arbitrary probability density functions (PDFs). This idea was further explored by Cobb and Shenoy [2006a], and they describe how mixtures of truncated exponentials can be estimated to approximate any PDF using an optimization procedure [Cobb *et al.* 2004], and applied to hybrid Bayesian networks with linear deterministic variables [Cobb and Shenoy 2005a] and also to hybrid Bayesian networks with non-linear deterministic variables [Cobb and Shenoy 2005b].

Other approaches to inference in hybrid Bayesian networks are approximate, which include dynamically discretizing the continuous variables (see, e.g., Kozlov and Koller [1997]), or which use sampling methods to compute approximate marginals (see, e.g., Gilks *et al.* [1996], Koller *et al.* [1999], Chang and Tian [2002], Gogate and Dechter [2005]).

In this paper, we explore an alternative strategy for exact inference in hybrid Bayesian networks. It is well known that mixtures of Gaussians can approximate any probability distribution [Titterington *et al.* 1985]. Thus, in principle, it should be possible to solve any hybrid Bayesian networks by first approximating such a network by a MoG Bayesian network, and then using the LJ algorithm to solve the MoG Bayesian network exactly. In this paper, we describe how this can be done. First we describe how non-Gaussian distributions can be approximated by a MoG distribution using an optimization method similar to that used by Cobb *et al.* [2004]. Second, we describe how to transform a network in which a discrete variable has continuous parents to a network where discrete variables do not have continuous parents using a series of arc reversals. Arc reversals were initially proposed by Olmsted [1983] and Shachter [1986] in the context of solving influence diagrams. Third, we show how a Bayesian network with conditionally deterministic nodes that are a nonlinear function of their parents can be approximated by a MoG Bayesian network. We approximate a nonlinear function by a piecewise linear function. Finally, we show how a BN with a conditional Gaussian distribution whose variance is a function of its parents can be approximated by a MoG BN.

An outline of the remainder of this paper is as follows. In section 2, we describe how a mixture of Gaussians can approximate a non-Gaussian distribution. In section 3, we describe how a Bayesian network in which some discrete variables have continuous variables can be converted to a MoG Bayesian network. In section 4, we describe how a Bayesian network with conditionally deterministic nodes that are a nonlinear function of their parents can be approximated by a MoG Bayesian network. In section 5, we describe how a Bayesian network with a conditional Gaussian distribution whose variance is a function of its parents can be approximated by a MoG Bayesian network. In section 6, we solve a small example to illustrate our technique. Finally in section 7, we conclude with a summary and some issues for future research.

# 2 NON-GAUSSIAN DISTRIBUTIONS

In this section, we will describe a methodology for approximating any non-Gaussian distribution by a MoG distribution. Poland and Shachter [1993] describe a method based on the EM algorithm [Dempster *et al.* 1977] for approximating continuous distributions by mixtures of Gaussians. Here we will describe an alternative method based on minimizing some measure of distance between two distributions to find a good approximation of a non-Gaussian distribution by a MoG distribution. We will illustrate our method for the uniform distribution over the unit interval [0, 1].

Let $A$ denote a chance variable that has the uniform distribution over the unit interval, denoted by $U[0, 1]$, and let $f_A$ denote its probability density function (PDF). Thus

$$f_A(x) = 1 \quad \text{if } 0 \le x \le 1$$
$$= 0 \quad \text{otherwise}$$

In approximating the PDF $f_A$ by a mixture of Gaussians, we first need to decide on the number of Gaussian components needed for an acceptable approximation. In this particular problem, more the components used, better will be the approximation. However, more components will lead to a bigger computational load in making inferences. We will measure the goodness of an approximation by estimating the Kullback-Liebler [1951] divergence measure between the target distribution and the corresponding MoG distribution.

Suppose we decide to use five components. Then we will approximate $f_A$ by the mixture PDF $g_A = p_1\,\varphi_{\mu_1,\sigma_1} + \ldots + p_5\,\varphi_{\mu_5,\sigma_5}$, where $\varphi_{\mu_i,\sigma_i}$ denote the PDF of a uni-variate Gaussian distribution with mean $\mu_i$ and standard deviation $\sigma_i > 0$, $p_1, \ldots, p_5 \ge 0$, and $p_1 + \ldots + p_5 = 1$. To estimate the mixture PDF, we need to estimate fourteen free parameters, e.g., $p_1, \ldots, p_4, \mu_1, \ldots, \mu_5, \sigma_1, \ldots, \sigma_5$. Based on the symmetry of $f_A$ around $a = 0.5$, we can reduce the number of free parameters to 7 by assuming that $p_1 = p_5$, $p_2 = p_4$, $\mu_3 = 0.5$, $\mu_4 = 1–\mu_2$, $\mu_5 = 1–\mu_1$, $\sigma_1 = \sigma_5$, and $\sigma_2 = \sigma_4$. To find the values of the 7 free parameters, we solve a non-linear optimization problem as follows:

Find $p_1, p_2, \mu_1, \mu_2, \sigma_1, \sigma_2, \sigma_3$ so as to minimize $\delta(f_A, g_A)$

subject to: $p_1 \ge 0, p_2 \ge 0, 2p_1+2p_2 \le 1, \sigma_1 \ge 0, \sigma_2 \ge 0, \sigma_3 \ge 0$,

where $\delta(f_A, g_A)$ denotes a distance measure between two PDFs. A commonly used distance measure that is easy to optimize is the sum of squared errors $\delta_{SSE}$ defined as follows:

$$\delta_{SSE}(f_A, g_A) = \int_S (f_A(x) - g_A(x))^2\, dx$$

In practice, we solve a discrete version of the non-linear optimization problem by discretizing both $f_A$ and $g_A$ using a large number of bins. To discretize $g_A$, we assume that the domain of $\varphi_{\mu_i,\sigma_i}$ extends only from $\mu_i–3\sigma_i$ to $\mu_i+3\sigma_i$. To match the domain of the $U[0, 1]$ distribution, we constrain the values $\mu_i–3\sigma_i \ge 0$, and $\mu_i+3\sigma_i \le 1$ for $i = 1, \ldots, 5$. Suppose we divide the domain [0, 1] into $n$ equally sized bins. Let $f_i$ and $g_i$ denote the probability masses for the $i^{th}$ bin corresponding to PDFs $f_A$ and $g_A$, respectively Then the discrete version of the non-linear programming problem can be stated as follows:

Minimize $\sum_{i=1}^{n} (f_i - g_i)^2$

subject to: $p_1 \ge 0, p_2 \ge 0, 2p_1+2p_2 \le 1, \sigma_1 \ge 0, \sigma_2 \ge 0, \sigma_3 \ge 0, \mu_1–3\sigma_1 \ge 0, \mu_2–3\sigma_2 \ge 0, 0.5 –3\sigma_3 \ge 0$.

One can use the solver in Excel to solve such optimization problems taking care to avoid local optimal solutions. An optimal solution computed in Excel with $n = 100$ (shown rounded to 3 digits) is as follows: $p_1 = p_5 = 0.169$, $p_2 = p_4 = 0.070$, $p_3 = 0.522$, $\mu_1 = 0.069$, $\mu_2 = 0.197$, $\mu_3 = 0.5$, $\sigma_1 = \sigma_5 = 0.023$, $\sigma_2 = \sigma_4 = 0.066$, $\sigma_3 = 0.167$. A graph of the two PDFs overlaid over each other is shown in Figure 1.

To measure the goodness of the approximation, we can compute the Kullback-Leibler (KL) divergence of the two distributions over the domain where both densities are positive, i.e., [0, 1]. The KL divergence is approximately 0.042. We can also compare moments. The mean and variance of the $U[0, 1]$ distribution are 0.5 and 0.083. The mean and variance of the MoG approximation are 0.5 and 0.073. We can also compare the cumulative distribution functions (CDF). A graph of the two CDFs overlaid over each other is shown in Figure 2.

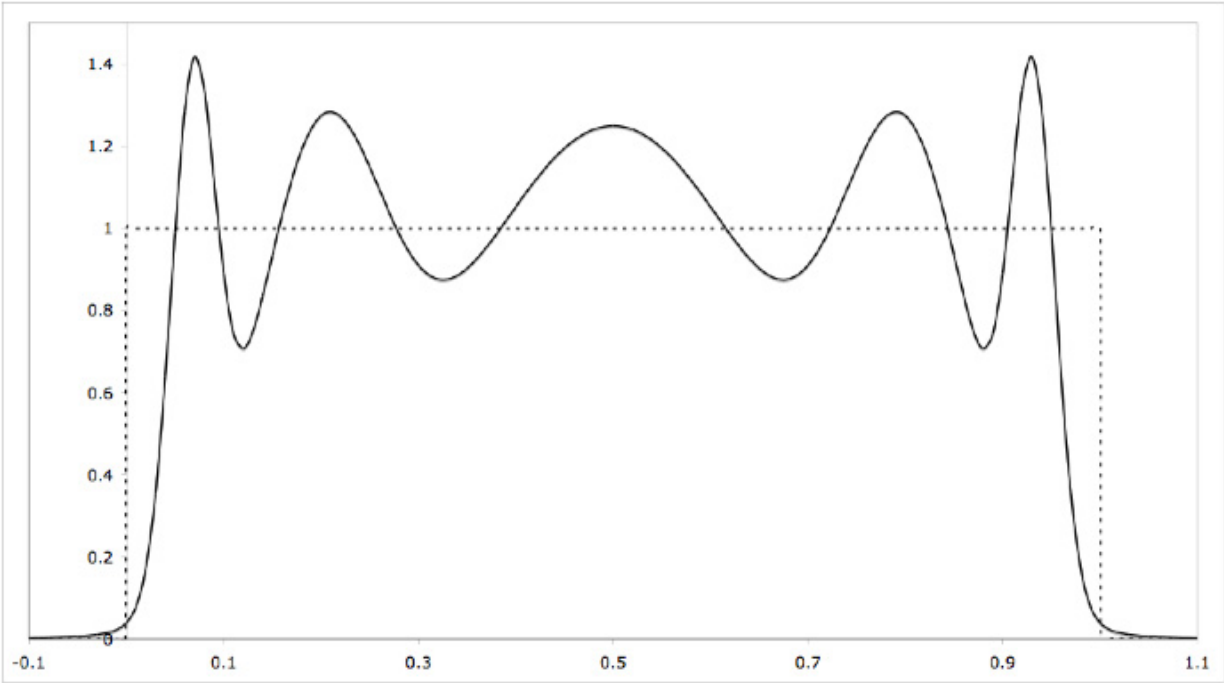

Figure 1: A 5-component MoG Approximation (solid) of the $U[0, 1]$ Distribution (dashed).

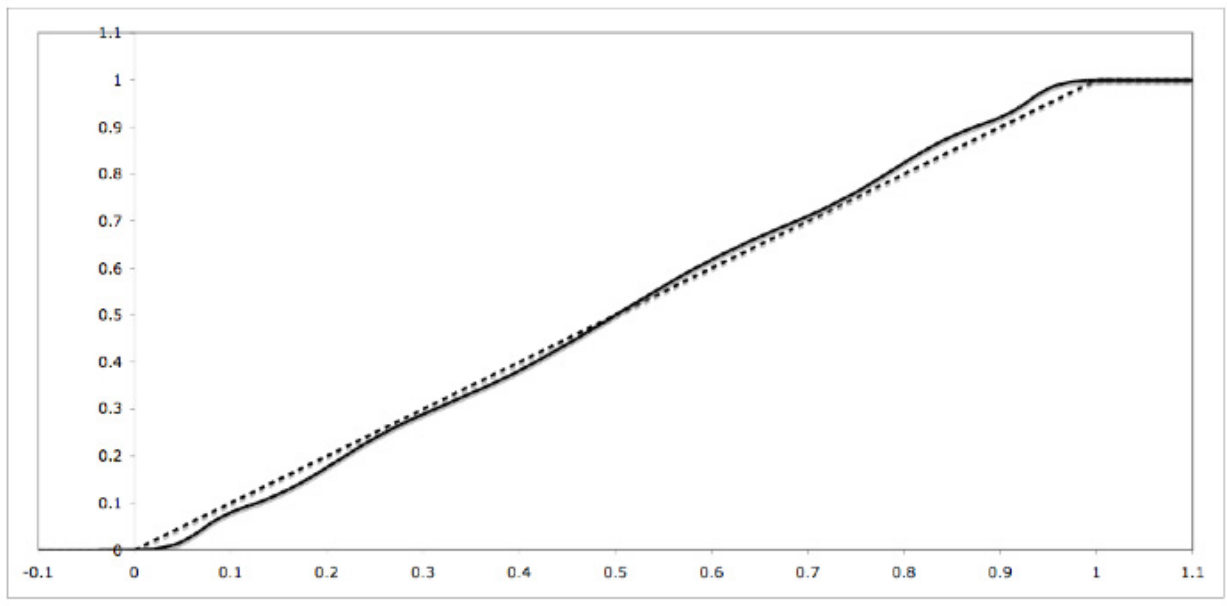

Figure 2: The CDFs of the $U[0, 1]$ Distribution (dashed), and its MoG Approximation (solid).

In a Bayesian network, chance node $A$ with the $U[0, 1]$ distribution is approximated by a Bayes net component that has a discrete variable $S$ with five states and a continuous variable $A$ whose conditional distributions are conditional linear Gaussians as shown in Figure 3. Borrowing the convention from Hugin, discrete nodes are depicted by single border ovals and continuous variables are denoted by double border ovals. The marginal distribution of $A$ is approximately uniform over the unit interval. The selector variable $S$ has no significance. It is just a synthetic variable whose only role is to provide weights for the mixture distribution.

If we wish to model a uniform distribution over the range $[a, b]$, denoted by $U[a, b]$, then one easy way is to introduce a conditionally deterministic variable $C$, which is a linear function of $A$ as follows: $C|a \sim N((b–a)A + a, 0)$ Thus, if $A \sim U[0, 1]$, then $C \sim U[a, b]$. Alternatively, we can find a mixture of Gaussians approximation of $U[a, b]$ directly, by having the selector variable $S$ with the same distribution as before, and changing the conditional linear Gaussian distributions of $A$ as follows: $A|s_1 \sim N((b–a)0.069+a, (b–a)^2 0.023^2)$, …, $A|s_5 \sim N((b–a)0.931+a, (b–a)^2 0.023^2)$. It is easy to confirm that the marginal distribution of $A$ is approximately $U[a, b]$.

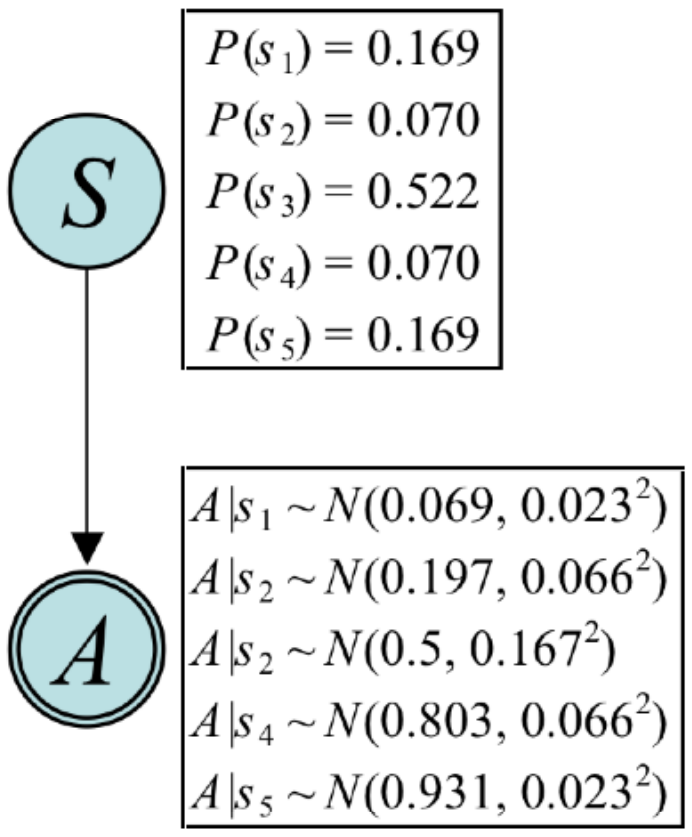

Figure 3: A MoG Bayesian Network Approximation of the $U[0, 1]$ Distribution.

## 3 DISCRETE NODES WITH CONTINUOUS PARENTS

MoG Bayesian networks do not allow discrete nodes with continuous parents. Often in the process of modeling situations, we are faced with a network in which some discrete nodes have continuous parents. In this section, we describe a method for approximating such Bayesian networks by MoG Bayesian networks.

Since MoG Bayesian networks do not allow discrete nodes with continuous parents, we change the topology of the network using arc reversals. Given a Bayesian network, we perform a series of arc reversals so that all discrete nodes precede continuous variables in any ordering that is compatible with the arcs in the Bayesian network (in the sense that if we have an arc from $X$ to $Y$, then $X$ precedes $Y$ in the ordering). After we have completed the arc reversals, we then proceed to approximate the conditional distributions of continuous nodes by mixtures of Gaussians using the technique described in the previous section. We will illustrate our proposed method using a simple example.

Consider a Bayesian network with two chance variables $A$ and $B$, and an arc from $A$ to $B$. $A$ is a continuous variable with the standard normal $N(0, 1)$ distribution, and $B$ is a binary discrete chance variable with states $b$ and $nb$, whose conditional distribution is given by the logistic function:

$$P(B = b \mid A = a) = 1/(1 + e^{-2a}), \text{ and}$$
$$P(B = nb \mid A = a) = e^{-2a}/(1 + e^{-2a}).$$

A graph of the logistic function is shown in Figure 4. Notice that the logistic function is symmetric about the axis $a = 0$ in the sense that $P(b \mid a) = 1 - P(b \mid -a)$ for $a > 0$. Also, $P(nb \mid a) = 1 - P(b \mid a) = P(b \mid -a)$.

Since the joint distribution of $A$ and $B$ is mixed (with probability masses and densities), we will describe it using the notation of mixed potentials introduced in Cobb and Shenoy [2005a]. The values of a mixed potential have two parts, mass and density. When we have only probability masses for discrete variables, we will use the symbol ι in the density part to denote an absence of density (or vacuous density), and it has the property that it is an identity for the semigroup of density potentials with the pointwise multiplication operation. Thus the potential $\alpha$ for $A$ is given by $\alpha(a) = (1, \varphi_{0,1}(a))$, where $\varphi_{0,1}(a)$ is the PDF of the standard normal distribution. The 1 in the mass part can be interpreted as a weight for the density value in the density part. The potential $\beta$ for $\{A, B\}$ representing the conditional distribution of $B$ given $A$ is given by

$$\beta(a, b) = (1/(1+e^{-2a}), \iota), \text{ and}$$

$$\beta(a, nb) = (e^{-2a}/(1+e^{-2a}), \iota).$$

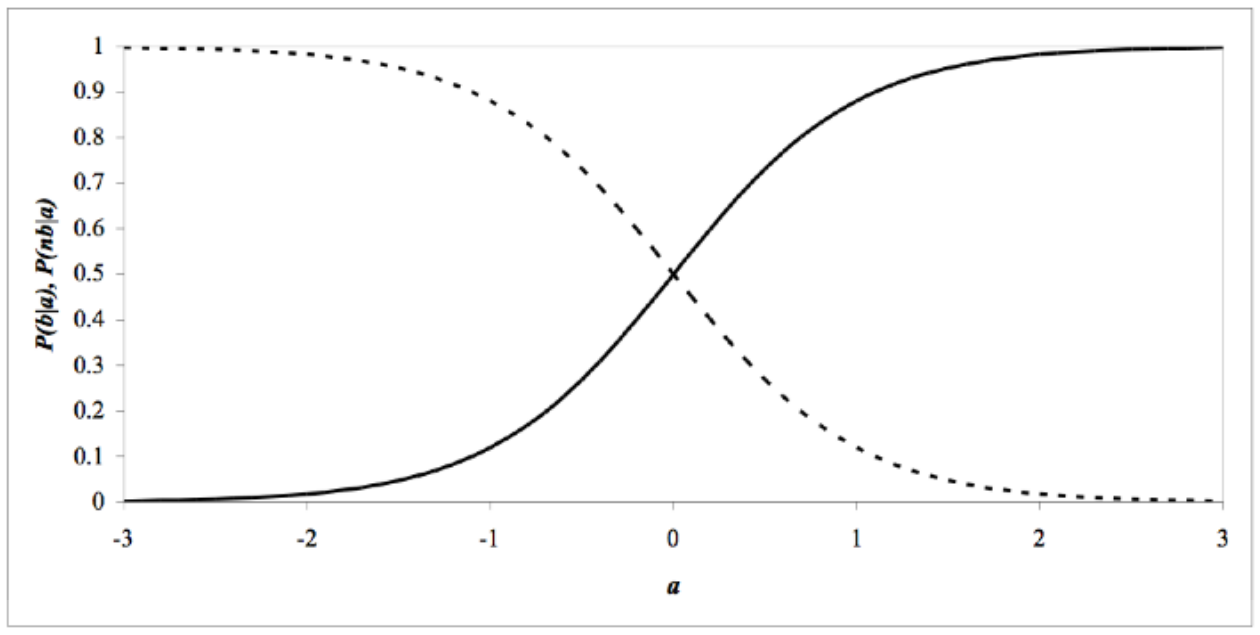

Figure 4: A Graph of the Conditional Probability $P(b \mid a)$ (solid) and $P(nb \mid a)$ (dashed).

To reverse the arc from $A$ to $B$, we perform the usual operations. First, we combine the mixed potentials associated with $A$ and $B$ using pointwise multiplication of the corresponding mass and density parts. The combination is denoted by $\alpha \otimes \beta$. Thus, $\alpha \otimes \beta$ is a potential for $\{A, B\}$ given by

$$(\alpha \otimes \beta)(a, b) = (1/(1+e^{-2a}), \varphi_{0,1}(a)), \text{ and}$$

$$(\alpha \otimes \beta)(a, nb) = (e^{-2a}/(1+e^{-2a}), \varphi_{0,1}(a)).$$

Next, we find the marginal for $B$ by marginalizing $A$ out of the joint potential $\alpha \otimes \beta$. Since $A$ is a continuous variable, it is marginalized out by integrating the product of the mass and density parts over the domain of $A$. We denote the marginal by $(\alpha \otimes \beta)^{\downarrow B}$ or $(\alpha \otimes \beta)^{-A}$, depending on whether we wish to emphasize the variables that remain, or the variables that are marginalized out. Thus,

$$(\alpha \otimes \beta)^{\downarrow B}(b) = \left( \int_{-\infty}^{\infty} (1/(1+e^{-2x})\varphi_{0,1}(x)dx, \iota \right) = (0.5, \iota)$$

$$(\alpha \otimes \beta)^{\downarrow B}(nb) = \left( \int_{-\infty}^{\infty} (e^{-2x}/(1+e^{-2x})\varphi_{0,1}(x)dx, \iota \right) = (0.5, \iota)$$

In general, we may have to do a numerical integration to do this operation if the integrals are not easily done in closed form. In our case, since both the logistic and density function are symmetric about the axis $a = 0$, the results can be deduced.

Next, to compute the conditional distributions for $A$ given $B$, we divide the joint potential $\alpha \otimes \beta$ by the marginal $(\alpha \otimes \beta)^{\downarrow B}$ using pointwise division. We denote the division by $(\alpha \otimes \beta) \oslash (\alpha \otimes \beta)^{\downarrow B}$. Thus,

$$((\alpha \otimes \beta) \oslash (\alpha \otimes \beta)^{\downarrow B})(a, b) = (2/(1+e^{-2a}), \varphi_{0,1}(a)) = (1, (2/(1+e^{-2a}))\varphi_{0,1}(a)), \text{ and}$$

$$((\alpha \otimes \beta) \oslash (\alpha \otimes \beta)^{\downarrow B})(a, nb) = (2e^{-2a}/(1+e^{-2a}), \varphi_{0,1}(a)) = (1, (2e^{-2a}/(1+e^{-2a}))\varphi_{0,1}(a)).$$

Finally, we need to approximate the conditional density functions $(2/(1+e^{-2a}))\varphi_{0,1}(a)$ and $(2e^{-2a}/(1+e^{-2a}))\varphi_{0,1}(a)$ by mixtures of Gaussians. To do this, we use the technique described in the previous section. We used only two components to approximate these density functions, and an approximation is given as follows (computed in Excel using $n = 600$, parameters rounded to 3 decimal places):

$$(2/(1+e^{-2a}))\varphi_{0,1}(a) \approx 0.070\ \varphi_{0.246,0.436} + 0.930\ \varphi_{0.613,0.796}$$

$$(2e^{-2a}/(1+e^{-2a}))\varphi_{0,1}(a) \approx 0.070\ \varphi_{-0.246,0.436} + 0.930\ \varphi_{-0.613,0.796}$$

A graph of the conditional density $f_{A|b}$ overlaid with the MoG approximation is shown in Figure 5.

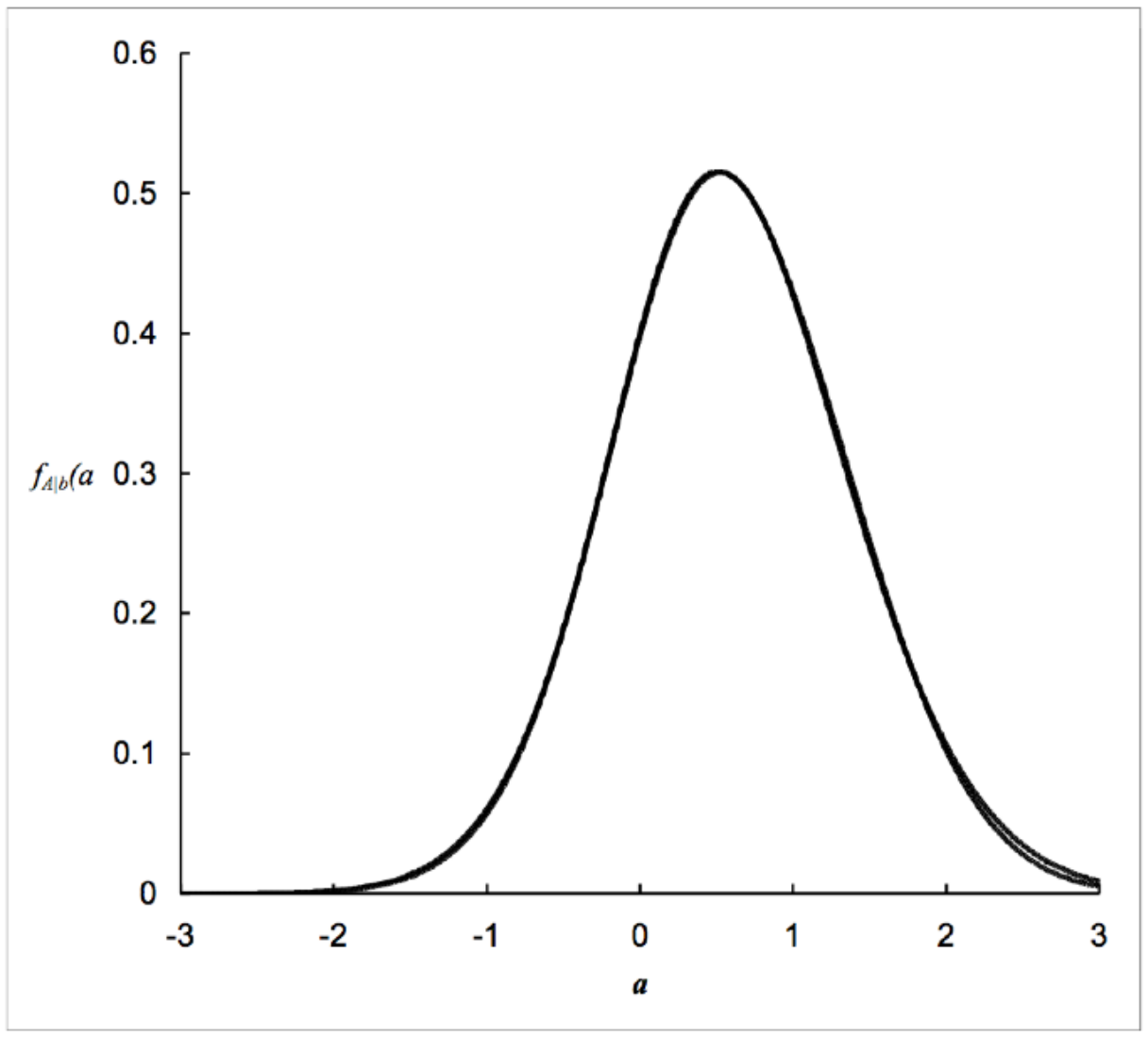

Figure 5: A Graph of the Conditional Density $f_{A|b}$ overlaid on its MoG Approximation.

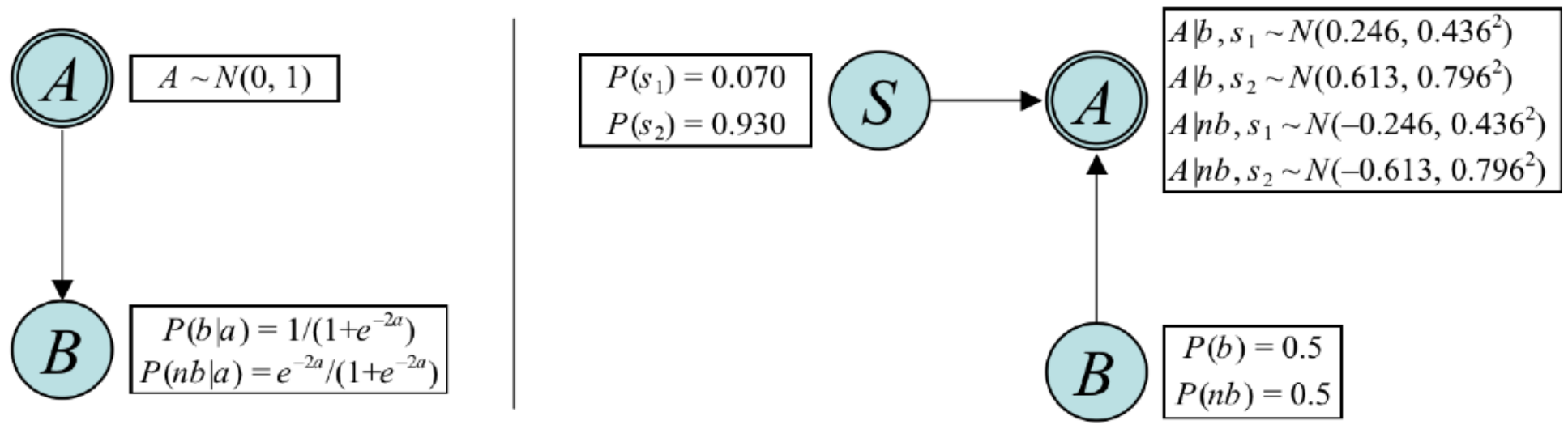

Figure 6: An Augmented CLG Bayesian Network and its MoG Approximation.

We have completed approximating the mixed Bayesian network by a mixture of Gaussians (MoG) Bayesian network. The original Bayesian network and its MoG approximation are shown in Figure 6.

To verify the quality of the approximation, we implemented the MoG approximation in Hugin and computed the following quantities. The marginal distribution of $A$ has mean 0 and variance 0.96 (compared to the exact mean 0 and variance 1). An observation, e.g., of $A$ = 1 yields marginal probability $P(B = b \mid A = 1)$ = 0.876 (compared to 0.881 given by the exact model).

## 4 NON-LINEAR CONDITIONALLY DETERMINISTIC NODES

In this section, we examine the problem of representing a Bayesian network with a non-linear conditionally deterministic variable as a MoG Bayesian network.

Consider a Bayesian network with two continuous variables $A$ and $B$ such that $A \sim N(0, 1)$, and $B|a \sim N(a^2, 0)$. Since the conditional distribution of $B$ is not linear Gaussian, this is not a MoG Bayesian network. The chance variable $B$ is a deterministic function of $A$, $B|a = a^2$ with probability 1. We say $B$ is "conditionally deterministic," and we depict conditionally deterministic variables with triple-bordered circles. A BN graph for this example is shown in Figure 8.

To approximate this Bayesian network by a MoG Bayesian network, we employ the same idea as in Cobb and Shenoy [2005b]. We approximate the non-linear deterministic function $B|a = a^2$ by a piecewise linear function. Suppose, e.g., we approximate the deterministic function $B|a = a^2$ by a six-segment piecewise linear approximation as follows:

$$B|a = \begin{cases} -5a - 6 & \text{for } -\infty \le a < -2 \\ -3a - 2 & \text{for } -2 \le a < -1 \\ -a & \text{for } -1 \le a < 0 \\ a & \text{for } 0 \le a < 1 \\ 3a - 2 & \text{for } 1 \le a < 2 \\ 5a - 6 & \text{for } 2 \le a \le \infty \end{cases}$$

A graph of the piecewise linear approximation appears overlaid on the actual function in Figure 7. Next, we approximate the Bayesian network with the network shown in Figure 8. In this Bayesian network, $S$ is a discrete indicator variable with six states corresponding to the six regions of the piecewise linear approximation. The conditional probability distribution for $S$ is as follows. If $a$ is in region $i$, then $P(S = s_i \mid a) = 1$, $P(S = s_j \mid a) = 0$ for $j \neq i$. The conditional probability distribution of $B$ given $a$ and $s_i$ is given by the linear approximation of $a^2$ in the region $s_i$.

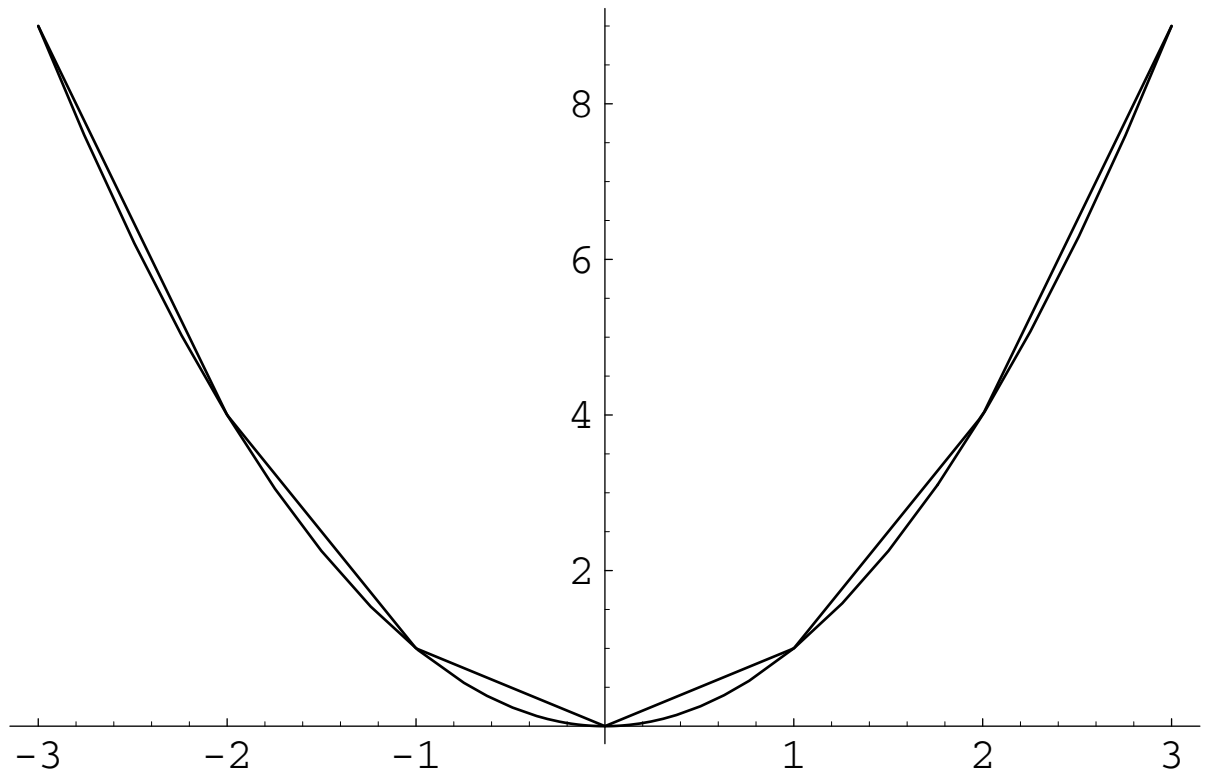

Figure 7: A Piecewise Linear Approximation of the Nonlinear Function.

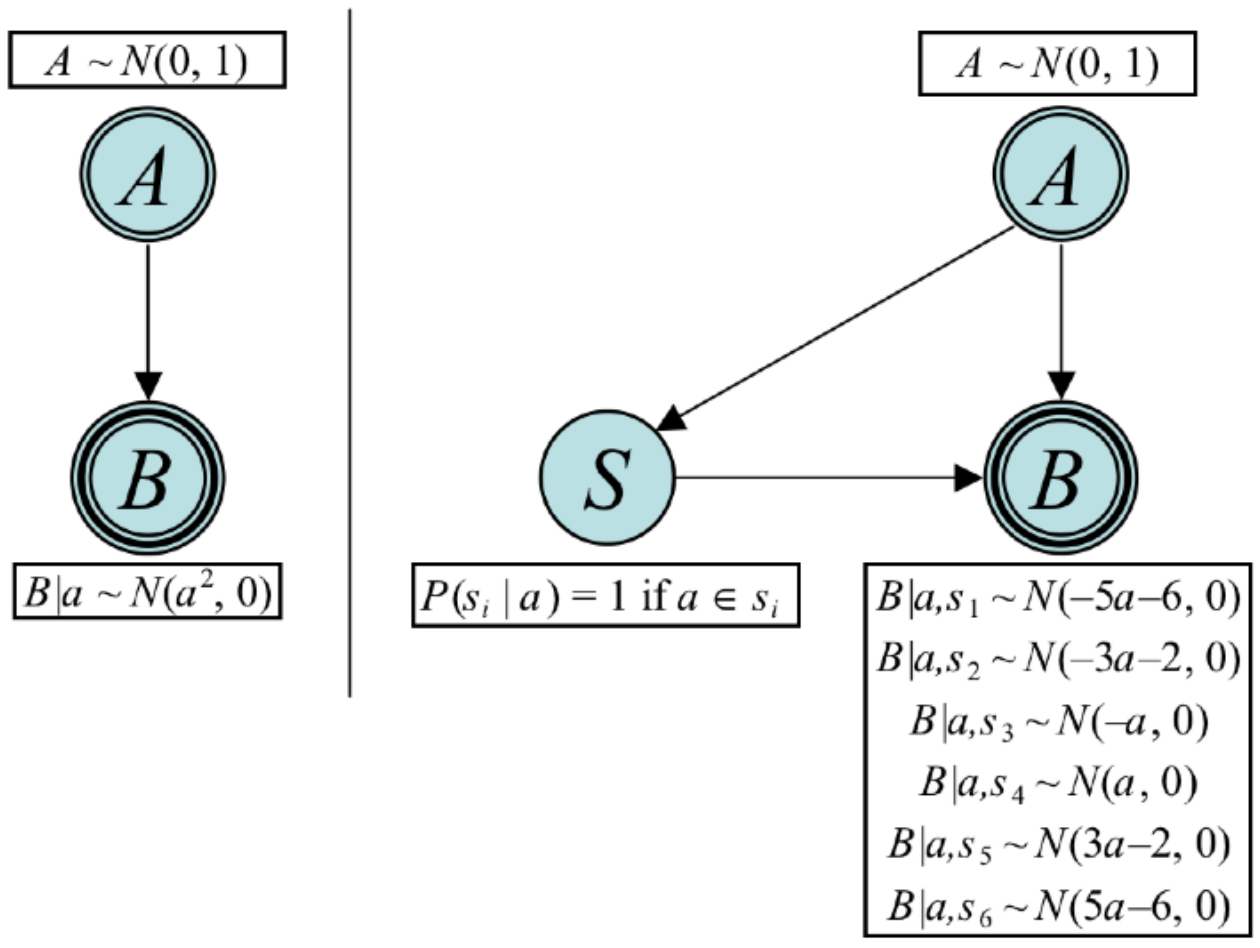

Figure 8: A Nonlinear Function Approximated by a Piecewise Linear Function.

| $\mu_{ij}$ | $s_1$ | $s_2$ | $s_3$ | $s_4$ | $s_5$ | $s_6$ |
|---|---|---|---|---|---|---|
| $r_1$ | -2.500 | -1.827 | -0.929 | 0.929 | 1.827 | 2.500 |
| $r_2$ | -2.235 | -1.500 | -0.799 | 0.799 | 1.500 | 2.235 |
| $r_3$ | -2.112 | -1.236 | -0.500 | 0.500 | 1.236 | 2.112 |
| $r_4$ | -2.050 | -1.112 | -0.218 | 0.218 | 1.112 | 2.050 |
| $r_5$ | -2.019 | -1.049 | -0.091 | 0.091 | 1.049 | 2.019 |
| $r_6$ | -2.003 | -1.017 | -0.031 | 0.031 | 1.017 | 2.003 |

| $P(r_i \mid s_j)$ | $s_1$ | $s_2$ | $s_3$ | $s_4$ | $s_5$ | $s_6$ |
|---|---|---|---|---|---|---|
| $r_1$ | 0.409 | 0.086 | 0.052 | 0.052 | 0.086 | 0.409 |
| $r_2$ | 0.274 | 0.459 | 0.133 | 0.133 | 0.459 | 0.274 |
| $r_3$ | 0.164 | 0.230 | 0.504 | 0.504 | 0.230 | 0.164 |
| $r_4$ | 0.088 | 0.128 | 0.187 | 0.187 | 0.128 | 0.088 |
| $r_5$ | 0.044 | 0.065 | 0.087 | 0.087 | 0.065 | 0.044 |
| $r_6$ | 0.021 | 0.032 | 0.036 | 0.036 | 0.032 | 0.021 |

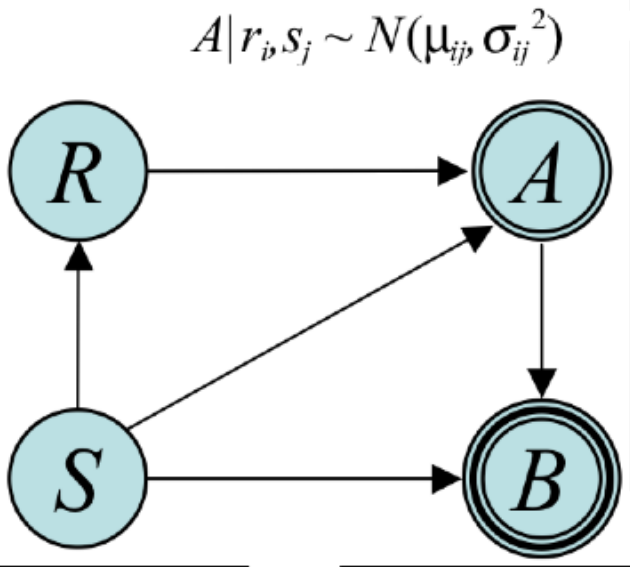

| $\sigma_{ij}$ | $s_1$ | $s_2$ | $s_3$ | $s_4$ | $s_5$ | $s_6$ |
|---|---|---|---|---|---|---|
| $r_1$ | 0.167 | 0.058 | 0.024 | 0.024 | 0.058 | 0.167 |
| $r_2$ | 0.078 | 0.167 | 0.067 | 0.067 | 0.167 | 0.078 |
| $r_3$ | 0.037 | 0.079 | 0.167 | 0.167 | 0.079 | 0.037 |
| $r_4$ | 0.017 | 0.037 | 0.073 | 0.073 | 0.037 | 0.017 |
| $r_5$ | 0.006 | 0.016 | 0.030 | 0.030 | 0.016 | 0.006 |
| $r_6$ | 0.001 | 0.006 | 0.010 | 0.010 | 0.006 | 0.001 |

| | |
|---|---|
| $P(s_1)$ = | 0.021 |
| $P(s_2)$ = | 0.136 |
| $P(s_3)$ = | 0.342 |
| $P(s_4)$ = | 0.342 |
| $P(s_5)$ = | 0.136 |
| $P(s_6)$ = | 0.021 |

| |
|---|
| $B \mid a, s_1 \sim N(-5a-6, 0)$ |
| $B \mid a, s_2 \sim N(-3a-2, 0)$ |
| $B \mid a, s_3 \sim N(-a, 0)$ |
| $B \mid a, s_4 \sim N(a, 0)$ |
| $B \mid a, s_5 \sim N(3a-2, 0)$ |
| $B \mid a, s_6 \sim N(5a-6, 0)$ |

Figure 9: A MoG Approximation of the BN in Figure 8.

The Bayesian network shown in the right-hand side of Figure 8 is not a MoG since $S$, which is discrete, has a continuous parent. We can approximate this BN by a MoG BN using the technique described in the previous section. The resulting MoG BN is shown in Figure 9. To verify the quality of the approximation this MoG BN was entered in Hugin. The mean of $B$ is reported as 1.12 and the variance is reported as 1.64. Since $B$ has a chi-square distribution with 1 degree of freedom, its true mean is 1 and its variance is 2.

# 5 NON-CONSTANT VARIANCE

In CLG distributions, the variance of the conditional Gaussian distribution is a constant. In this section, we examine the case where a continuous node has a Gaussian distribution with a variance that is a function of its parents.

Consider a Bayesian network with two continuous nodes $A$ and $B$, such that $A \sim N(3, 1)$ and $B|a \sim N(a, a^2)$. Notice that although the conditional distributions of $B$ are Gaussian, the marginal distribution of $B$ is not Gaussian. A graph of the probability density function of $B$ is shown in Figure 10. The exact mean of $B$ is 3 (= $E[E[B|A]] = E[A]$), the exact variance is 11 (= $E[Var[B|A]] + Var[E[B|A]] = E[A^2] + Var[A] = 2Var[A] + (E[A])^2$), and the mode is approximately 1.91.

To approximate the joint distribution of $A$ and $B$ with a MoG distribution, we divide the domain of $A$ into many small segments, and in each segment, we approximate the conditional distribution of $B$ by a CLG distribution. We introduce a discrete variable $S$ with state space $\{s_1, \ldots, s_6\}$. The variable $S$ indicates which segment $A$ lies in. Suppose we partition the domain of $A$ into six segments as follows $(-\infty, 1)$, $[1, 2)$, …, $[5, \infty)$. Thus, the conditional probability distribution for $S$ is as follows. If $a$ is in segment $i$, then $P(S = s_i \mid a) = 1$, $P(S = s_j \mid a) = 0$ for $j \neq i$. Next, we approximate the conditional distribution of $B$ (with $A$ and $S$ as its parents) with a CLG distribution as follows: $B|a, s_1 \sim N(a, 0.5^2)$, $B|a, s_2 \sim N(a, 1.5^2)$, $B|a, s_3 \sim N(a, 2.5^2)$, $B|a, s_4 \sim N(a, 3.5^2)$, $B|a, s_5 \sim N(a, 4.5^2)$, $B|a, s_6 \sim N(a, 5.5^2)$. The resulting BN, shown in Figure 11, is not a MoG since discrete variable $S$ has continuous variable $A$ as its parent. We reverse the arc $A \rightarrow S$ as we did in the previous section resulting in the MoG BN shown in Figure 12.

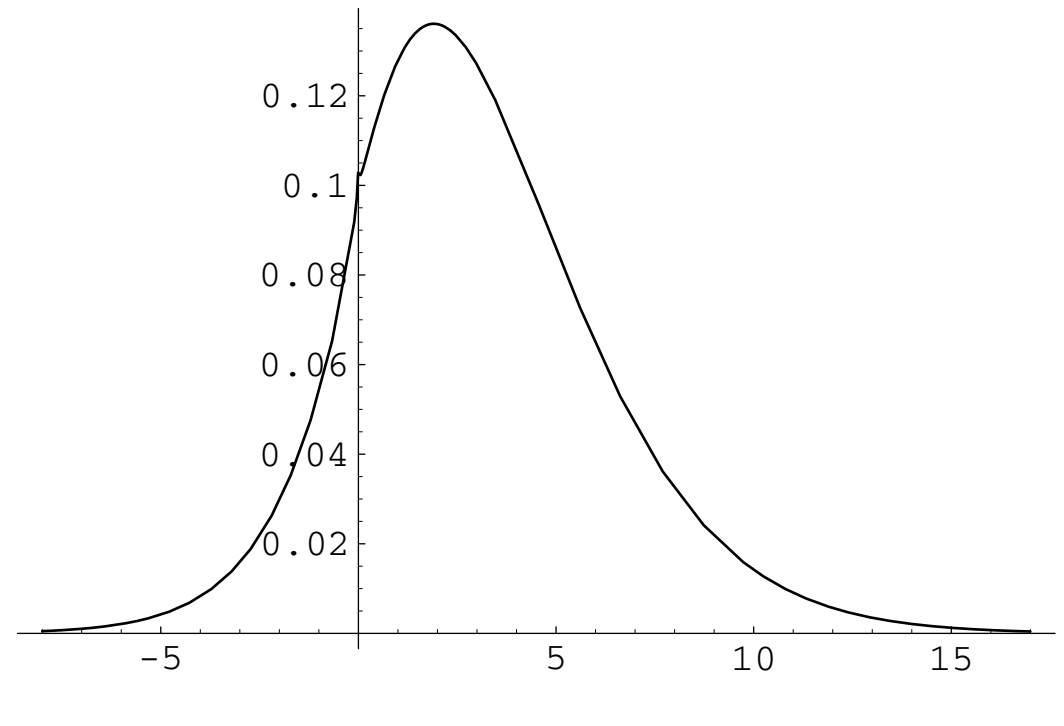

Figure 10: The Probability Density Function of $B$.

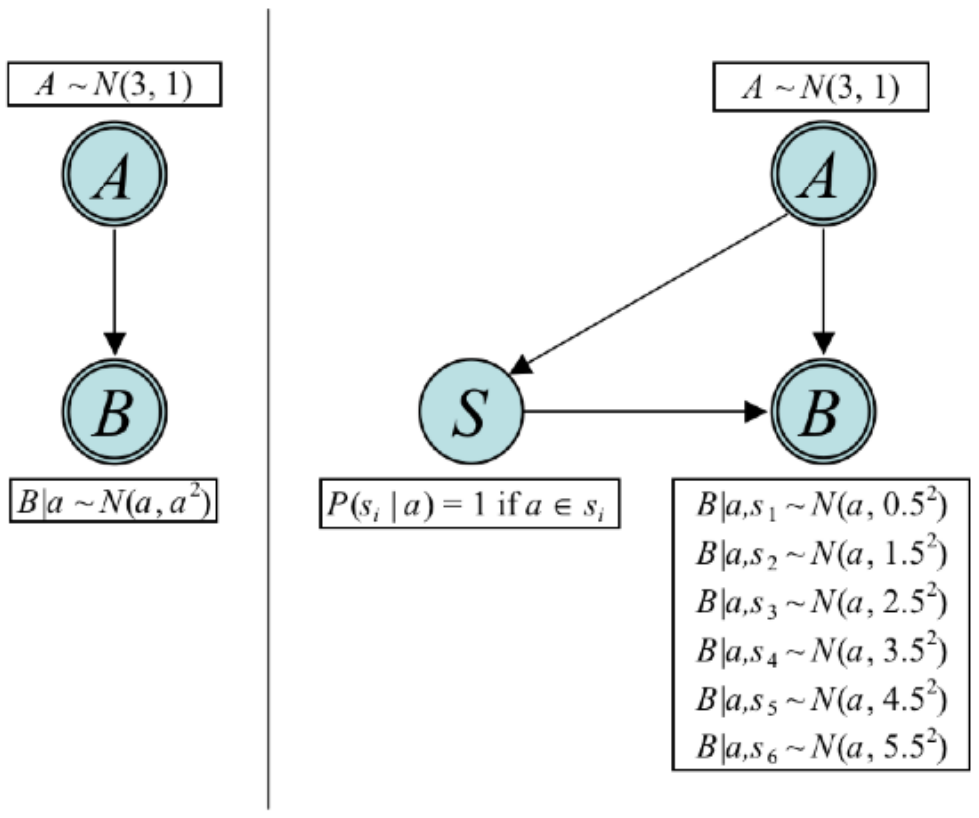

Figure 11: A Non-Constant Variance Approximated with a Constant Variance.

If we model this MoG BN in Hugin, the mean of the marginal distribution of $B$ is reported as 3 and the variance is reported as 11.004 (compared to the exact values 3 and 11, respectively).

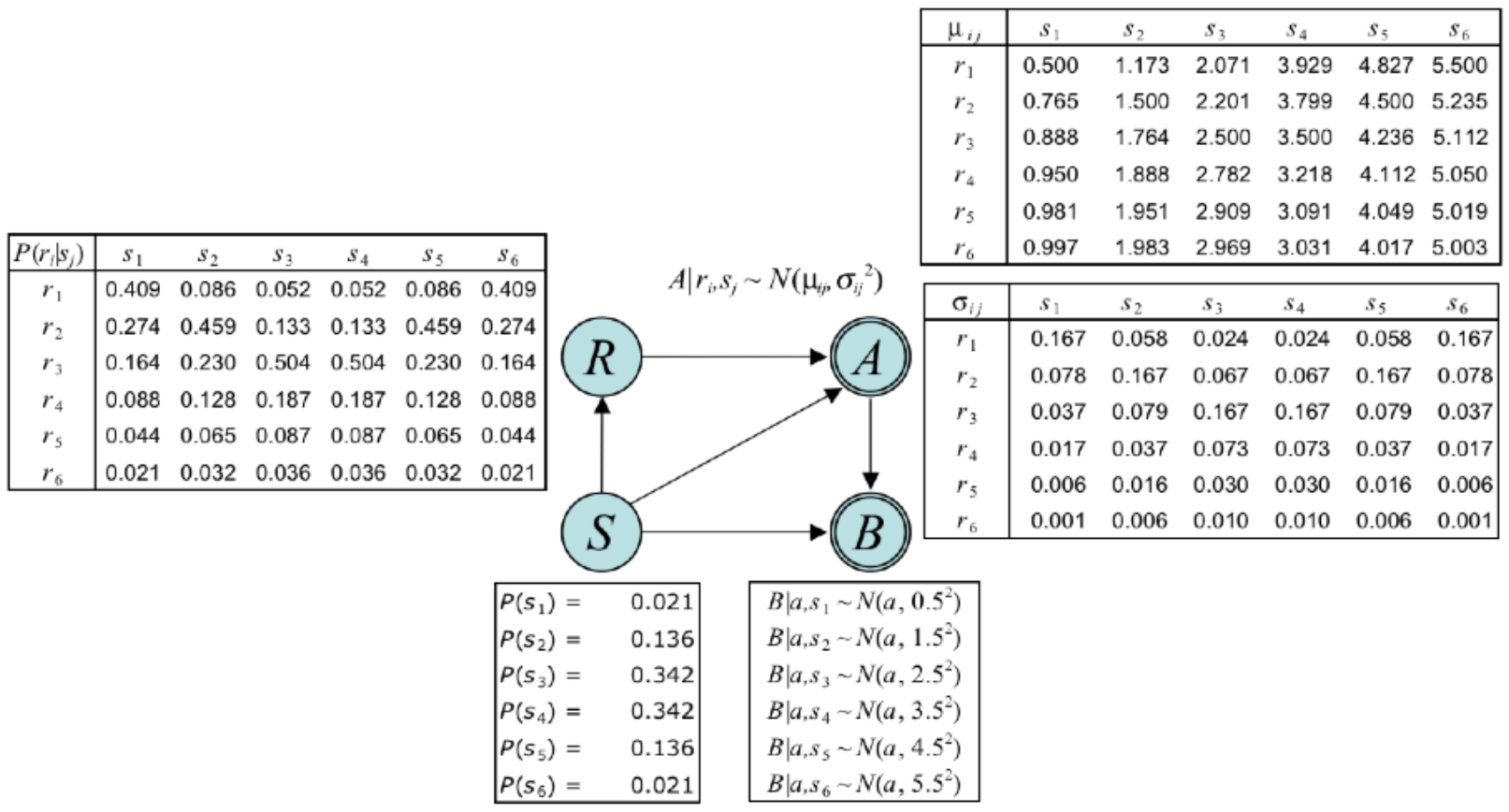

| $P(r_i \vert s_j)$ | $s_1$ | $s_2$ | $s_3$ | $s_4$ | $s_5$ | $s_6$ |
|---|---|---|---|---|---|---|
| $r_1$ | 0.409 | 0.086 | 0.052 | 0.052 | 0.086 | 0.409 |
| $r_2$ | 0.274 | 0.459 | 0.133 | 0.133 | 0.459 | 0.274 |
| $r_3$ | 0.164 | 0.230 | 0.504 | 0.504 | 0.230 | 0.164 |
| $r_4$ | 0.088 | 0.128 | 0.187 | 0.187 | 0.128 | 0.088 |
| $r_5$ | 0.044 | 0.065 | 0.087 | 0.087 | 0.065 | 0.044 |
| $r_6$ | 0.021 | 0.032 | 0.036 | 0.036 | 0.032 | 0.021 |

| $\mu_{ij}$ | $s_1$ | $s_2$ | $s_3$ | $s_4$ | $s_5$ | $s_6$ |
|---|---|---|---|---|---|---|
| $r_1$ | 0.500 | 1.173 | 2.071 | 3.929 | 4.827 | 5.500 |
| $r_2$ | 0.765 | 1.500 | 2.201 | 3.799 | 4.500 | 5.235 |
| $r_3$ | 0.888 | 1.764 | 2.500 | 3.500 | 4.236 | 5.112 |
| $r_4$ | 0.950 | 1.888 | 2.782 | 3.218 | 4.112 | 5.050 |
| $r_5$ | 0.981 | 1.951 | 2.909 | 3.091 | 4.049 | 5.019 |
| $r_6$ | 0.997 | 1.983 | 2.969 | 3.031 | 4.017 | 5.003 |

| $\sigma_{ij}$ | $s_1$ | $s_2$ | $s_3$ | $s_4$ | $s_5$ | $s_6$ |
|---|---|---|---|---|---|---|
| $r_1$ | 0.167 | 0.058 | 0.024 | 0.024 | 0.058 | 0.167 |
| $r_2$ | 0.078 | 0.167 | 0.067 | 0.067 | 0.167 | 0.078 |
| $r_3$ | 0.037 | 0.079 | 0.167 | 0.167 | 0.079 | 0.037 |
| $r_4$ | 0.017 | 0.037 | 0.073 | 0.073 | 0.037 | 0.017 |
| $r_5$ | 0.006 | 0.016 | 0.030 | 0.030 | 0.016 | 0.006 |
| $r_6$ | 0.001 | 0.006 | 0.010 | 0.010 | 0.006 | 0.001 |

Figure 12: A MoG Approximation of the BN in Figure 11.

# 6 AN EXAMPLE

In this section, we describe a slightly more complicated example with a mix of discrete, continuous, and conditionally deterministic variables. To approximate the non-MoG Bayesian network with a MoG Bayesian network, we need to do a series of arc reversals and approximate the resulting non-Gaussian PDFs with MoGs.

Consider a hybrid Bayesian network consisting of four variables as shown in Figure 13. This BN has two discrete and two continuous variables. One of the continuous variables is conditionally deterministic given its parent. Although the distributions of the continuous variables are conditional linear Gaussian, the BN is not MoG since the discrete variable $D$ has a continuous parent. This BN is an example of an augmented CLG model studied by Murphy [1999] and Lerner *et al*. [2001].

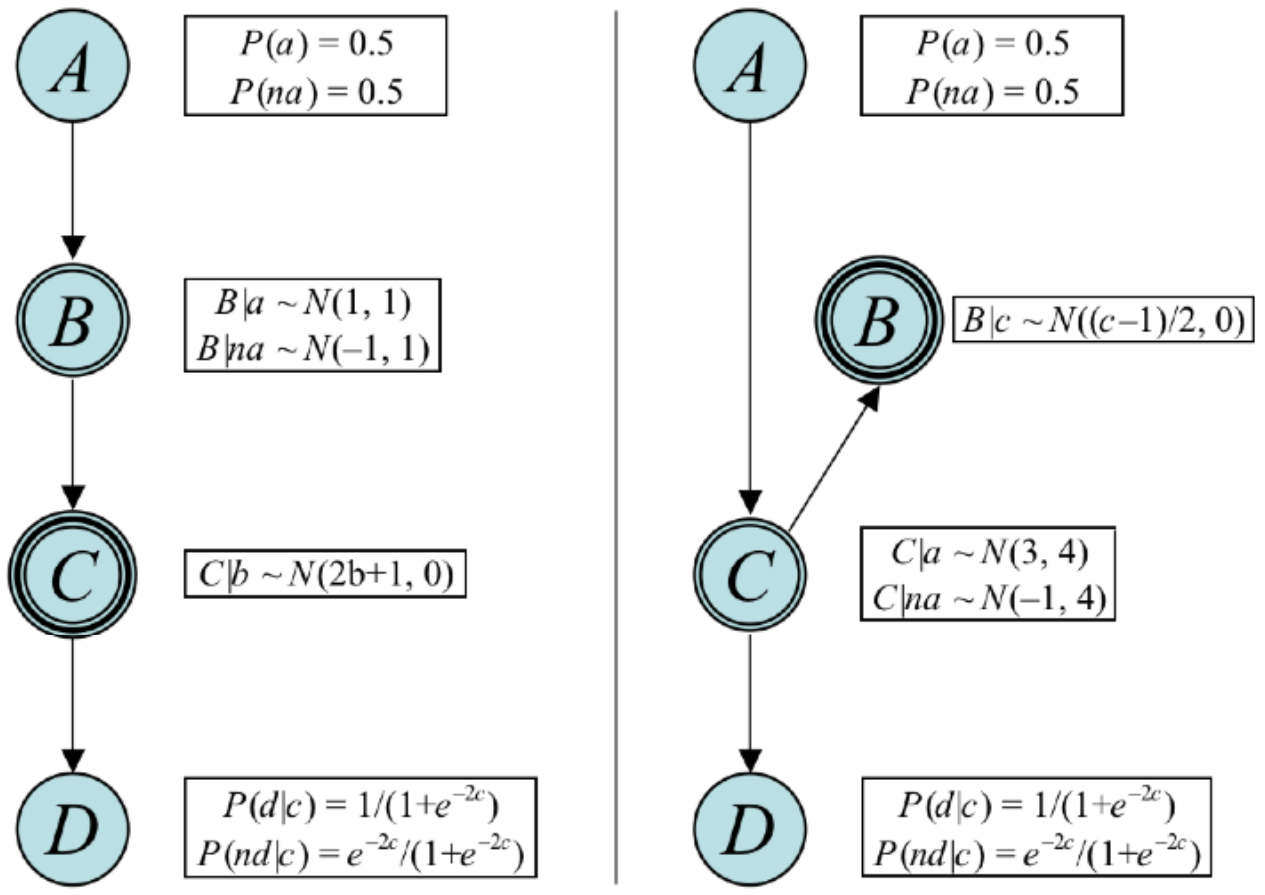

Figure 13: Left: A Non-MoG Bayesian network. Right: After Reversal of Arc ($B$, $C$).

## 6.1 ARC REVERSALS

A few words about arc reversals. First, a necessary condition for an arc to be reversible is that the two variables defining the arc have to be "adjacent." Two variables are said to be adjacent if there is a sequence of variables compatible with the arcs in a Bayes net (in the sense described earlier) such that the two variables are adjacent. Thus in the Bayes net in Figure 13, the following pairs of variables are adjacent: {$A$, $B$}, {$B$, $C$}, {$C$, $D$}. Second, an arc with a continuous conditionally deterministic variable at its tail cannot be reversed. This is because one of the operations involved in arc reversal is the computation of the conditional joint distribution (conditional on the union of the parents of the two variables) of the two variables defining the arc. This operation cannot be performed when we have only an equation defining the conditional distribution for the deterministic variable. Thus, in our example, the arc ($C$, $D$) cannot be reversed. An arc between two adjacent variables such that the variable at the head of the arc is deterministic can be reversed. The mechanics for reversing such arcs are slightly different from the usual operations for arc reversal [Cobb and Shenoy 2006b]. Consider the arc ($B$, $C$) in which $B$ is a continuous variable and $C$ is a continuous conditionally deterministic variable. The conditional joint density for {$B$, $C$} (given $A$) does not exist. However the conditional marginal density for $C$ (given $A$) does exist. Also, the conditional distribution for $B$ given $C$ (and $A$) is not a density function, but a deterministic function given by the inverse of the original deterministic function defining *C*.

In our example, after arc reversal, the distributions of *B* and *C* are as shown in Figure 13. Notice that after the arc reversal, *C* is a continuous variable and *B* is a continuous conditionally deterministic variable.

In the new BN, *C* and *D* are still adjacent and the arc (*C*, *D*) can be reversed. After reversing this arc, we get a BN as shown in Figure 14. Notice that the resulting BN has discrete variables that do not have continuous parents. Thus, this BN is almost MoG, except for the conditional distribution of variable *C*, which is not CLG. Of course, we can approximate the four conditional distributions by MoG as discussed in Section 2. We skip the details.

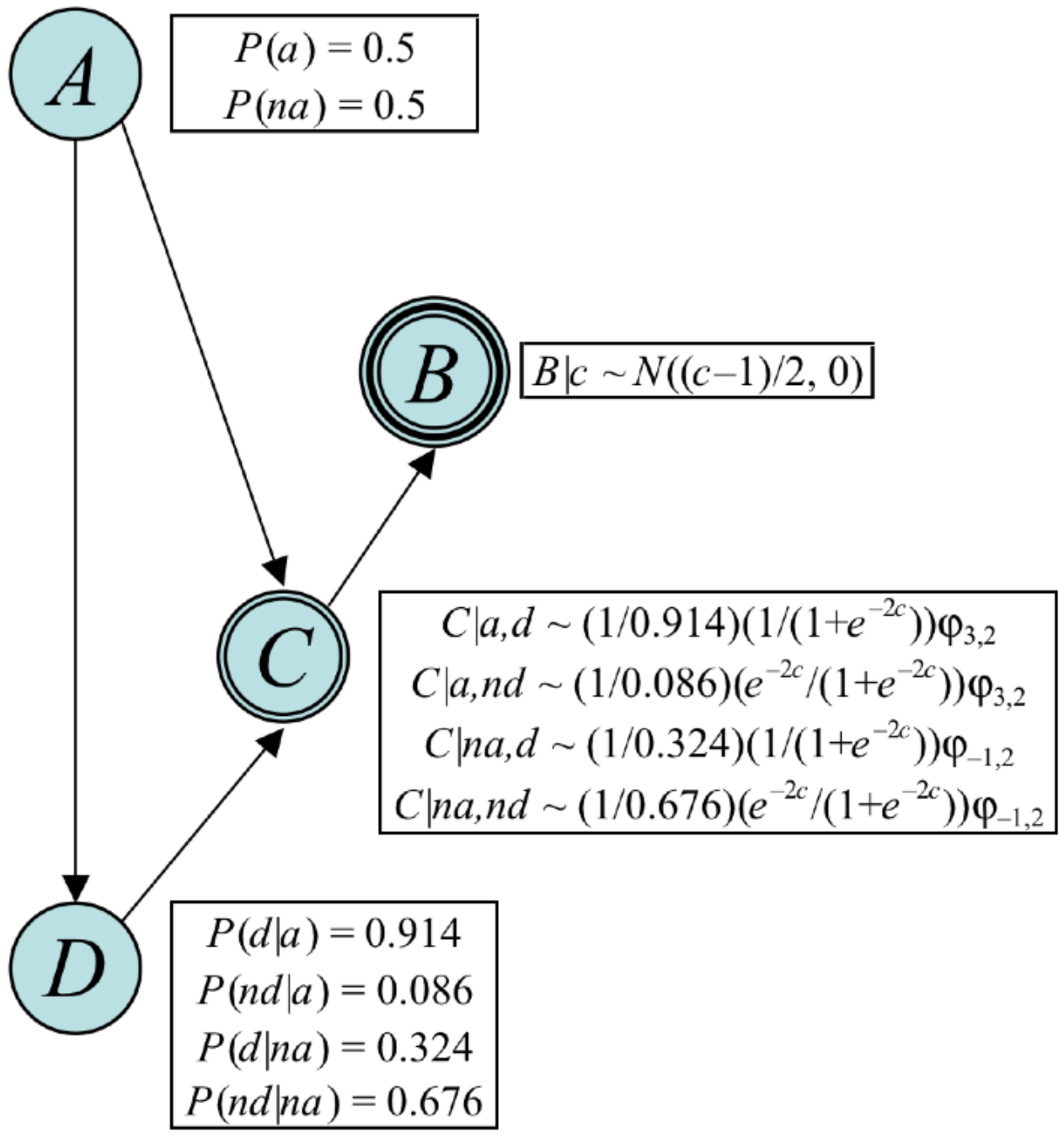

Figure 14: The BN Obtained from the BN in Figure 13 Right After Reversing Arc (*C*, *D*).

## 7 SUMMARY AND CONCLUSIONS

We have described a new method for "exact" inference in general hybrid BNs. Our method consists of first approximating a given hybrid BN by a MoG BN, and then using the LJ method for exact inference in MoG BNs.

A general hybrid BN can fail to be a MoG BN for many reasons. The conditional distribution of a continuous variable may not be a CLG distribution. There may be a discrete variable with continuous parents. A continuous variable may be a nonlinear deterministic function of its parents. The variance of a continuous variable may be a function of its parents. In this paper, we show some strategies for approximating non-MoG BNs by MoG BNs. In particular, we describe a general methodology for approximating an unconditional non-Gaussian distribution by a MoG distribution. We describe how we can transform a BN that has discrete variables with continuous parents to a MoG BN. We describe how we can approximate a BN with a nonlinear deterministic variable by a MoG BN. Finally, we describe how we can approximate a BN that has a Gaussian distribution with a non-constant variance by a MoG BN.

Our strategy of approximating non-MoG hybrid BNs by MoG BNs is based on the existence of a fast exact algorithm for making inferences in MoG BNs. However, the problem of making exact (or approximate) inferences in hybrid Bayes nets has been shown to be NP hard [Lerner and Parr 2001]. So there are no guarantees that our strategy will always work (i.e., while we may be successful in converting a hybrid BN to a MoG BN, the LJ algorithm may fail to compute marginals). But we are hopeful that our strategy will work for a broad class of problems. We expect the process of approximation to be done off line. And there exists a commercial implementation of the LJ algorithm that makes the use of our strategy practical.

The process of approximating a non-MoG BN by a MoG BN is not without costs. First, we add some dummy discrete variables that are not present in the original non-MoG BN. This, of course, adds to the computational load of making inferences in the resulting MoG BN. Second, in the process of reversing arcs, we increase the domains of potentials. This again increases the computational burden of making inferences in the MoG approximation. Finally, although the inference is exact, the quality of the approximation depends on how many components are used to approximate non-MoG BNs by MoG BNs. Further research is needed to quantify the quality of the approximation as a function of components used.

**Acknowledgements**

The inspiration for this paper came from a brief conversation I had with Steffen L. Lauritzen at UAI-03 in Acapulco, Mexico. This research builds on the stream of research done with Barry R. Cobb on the use of mixtures of truncated exponentials for inference in hybrid Bayesian networks. I am grateful to Mohammad M. Rahman for assisting me with the approximation of non-Gaussian densities with mixture of Gaussians, and to Barry Cobb for assisting me with the approximation of nonlinear conditionally deterministic variables.